%% file: paper_arxiv.tex
\documentclass[10pt,twocolumn,letterpaper]{article}

\usepackage{iccv}
\usepackage{times}
\usepackage{epsfig}
\usepackage{graphicx,subcaption}
\usepackage{amsmath}
\usepackage{amssymb}
\usepackage[table]{xcolor}
\usepackage{array}
\usepackage{dblfloatfix}
\usepackage{verbatim}
\usepackage{booktabs}
\usepackage{breqn}

\usepackage[breaklinks=true,bookmarks=false]{hyperref}

\iccvfinalcopy %

\def\iccvPaperID{2780} %

\ificcvfinal\fi
\newcommand{\datasetname}{DiDeMo}
\newcommand{\modelname}{MCN}
\DeclareMathOperator*{\argmin}{\arg\!\min}
\newcolumntype{L}{>{\centering\arraybackslash}m{2.2cm}}
\newcolumntype{M}{>{\centering\arraybackslash}m{2.2cm}}
\newcolumntype{N}{>{\centering\arraybackslash}m{0.5cm}}

\newcommand{\myparagraph}[1]{\vspace{3pt}\noindent{\bf #1}}
\begin{document}

\title{Localizing Moments in Video with Natural Language}
\renewcommand*{\thefootnote}{\fnsymbol{footnote}}
\author{Lisa Anne Hendricks$^{1\thanks{Work done at Adobe Research during LAH's summer internship}}$, Oliver Wang$^2$, Eli Shechtman$^2$, Josef Sivic$^{2,3\footnotemark[1]}$, Trevor Darrell$^1$, Bryan Russell$^2$\\
$^1$UC Berkeley, $^2$Adobe Research, $^3$INRIA\\
{\tt\small \url{https://people.eecs.berkeley.edu/~lisa_anne/didemo.html}}
}
\renewcommand*{\thefootnote}{\arabic{footnote}}

\maketitle
\begin{abstract}
We consider retrieving a specific temporal segment, or moment, from a video given a natural language text description.  
Methods designed to retrieve whole video clips with natural language determine what occurs in a video but not when.
To address this issue, we  propose the Moment Context Network (\modelname{}) which effectively localizes natural language queries in videos by integrating local and global video features over time.
A key obstacle to training our \modelname{} model is that current video datasets do not include pairs of localized video segments and referring expressions, or text descriptions which uniquely identify a corresponding moment.
Therefore, we collect the Distinct Describable Moments (\datasetname{}) dataset which consists of over 10,000 unedited, personal videos in diverse visual settings with pairs of localized video segments and referring expressions.
We demonstrate that \modelname{} outperforms several baseline methods and believe that our initial results together with the release of \datasetname{} will inspire further research on localizing video moments with natural language.
\end{abstract}

\input{introduction}
\input{related}

\input{method}

\input{dataset}

\input{results}
\section*{Acknowledgements}
TD was supported by DARPA, AFRL, DoD MURI award N000141110688, NSF awards IIS-1427425, IIS-1212798, and the Berkeley AI Research (BAIR) Lab.

\appendix
\section*{Supplemental}
\input{appendix}

\clearpage
{\small
\bibliographystyle{ieee}
\bibliography{bibliolong,egbib}
}

\end{document}

%% file: introduction.tex
\section{Introduction}

Consider the video depicted in Figure~\ref{fig:ConceptFigure}, in which a little girl jumps around, falls down, and then gets back up to start jumping again.
Suppose we want to refer to a particular temporal segment, or moment, from the video, such as when the girl resiliently begins jumping again after she has fallen. 
Simply referring to the moment via an action, object, or attribute keyword may not uniquely identify it.
For example, important objects in the scene, such as the girl, are present in each frame.
Likewise, recognizing all the frames in which the girl is jumping will not localize the moment of interest as the girl jumps both before and after she has fallen.
Rather than being defined by a single object or activity, the moment may be defined by when and how specific actions take place \emph{in relation} to other actions.
An intuitive way to refer to the moment is via a natural language phrase, such as ``the little girl jumps back up after falling''.

Motivated by this example, we consider localizing moments in video with natural language.
Specifically, given a video and text description, we identify start and end points in the video which correspond to the given text description.
This is a challenging task requiring both language and video understanding, with important applications in video retrieval, such as finding particular moments from a long personal holiday video, or desired B-roll stock video footage from a large video library (e.g., Adobe Stock\footnote{https://stock.adobe.com}, Getty\footnote{http://www.gettyimages.com}, Shutterstock\footnote{https://www.shutterstock.com}). 

Existing methods for natural language based video retrieval \cite{otani2016learning,xu2015jointly,torabi2016learning} retrieve an entire video given a text string but do not identify \textit{when} a moment occurs within a video.
To localize moments within a video we propose to learn a joint video-language model in which referring expressions and video features from corresponding moments are close in a shared embedding space.
However, in contrast to whole video retrieval, we argue that in addition to video features from a specific moment, global video context and knowing when a moment occurs within a longer video are important cues for moment retrieval.
For example, consider the text query ``The man on the stage comes closest to the audience''.
The term ``closest'' is relative and requires temporal context to properly comprehend.
Additionally, the temporal position of a moment in a longer video can help localize the moment.
For the text query ``The biker starts the race'', we expect moments earlier in the video in which the biker is racing to be closer to the text query than moments at the end of the video.
We thus propose the Moment Context Network (\modelname{}) which includes a global video feature to provide temporal context and a temporal endpoint feature to indicate when a moment occurs in a video.

A major obstacle when training our model is that current video-language datasets do not include natural language which can uniquely localize a moment.
Additionally, datasets like \cite{lin2014visual,regneri2013grounding}  are small and restricted to specific domains, such as dash-cam or cooking videos, while datasets \cite{chen2011collecting,rohrbach2015dataset,torabi2015using,xu2016msr-vtt} sourced from movies and YouTube are frequently edited and tend to only include entertaining moments (see \cite{sigurdsson2016hollywood} for discussion).
We believe the task of localizing moments with natural language is particularly interesting in unedited videos which tend to include uneventful video segments that would generally be cut from edited videos.
Consequently, we desire a dataset which consists of distinct moments from unedited video footage paired with descriptions which can uniquely localize each moment, analogous to datasets that pair distinct image regions with descriptions \cite{kazemzadeh2014referitgame,mao2015generation}.

To address this problem, we collect the Distinct Describable Moments (\datasetname{}) dataset which includes distinct video moments paired with descriptions which uniquely localize the moment in the video.
Our dataset consists of over 10,000 unedited videos with 3-5 pairs of descriptions and distinct moments per video.
\datasetname{} is collected in an open-world setting and includes diverse content such as pets, concerts, and sports games.
To ensure that descriptions are referring and thus uniquely localize a moment, we include a validation step inspired by \cite{kazemzadeh2014referitgame}.

\myparagraph{Contributions.} 
We consider the problem of localizing moments in video with natural language in a challenging open-world setting.  %
We propose the Moment Context Network (\modelname{}) which relies on local and global video features.
To train and evaluate our model, we collect the Distinct Describable Moments (\datasetname{}) dataset which consists of over 40,000 pairs of referring descriptions and localized moments in unedited videos.

%% file: related.tex
\section{Related Work}

Localizing moments in video with natural language is related to other vision tasks including video retrieval, video summarization, video description and question answering, and natural language object retrieval.
Though large scale datasets have been collected for each of these tasks, none fit the specific requirements needed to learn how to localize moments in video with natural language.

\myparagraph{Video Retrieval with Natural Language.}
Natural language video retrieval methods aim to retrieve a specific video given a natural language query.
Current methods \cite{otani2016learning,torabi2016learning,xu2015jointly} incorporate deep video-language embeddings similar to image-language embeddings proposed by \cite{frome2013devise,socher2014grounded}. %
Our method also relies on a joint video-language embedding.
However, to identify when events occur in a video, our video representation integrates local and global video features as well as temporal endpoint features which indicate when a candidate moment occurs within a video.

Some work has studied retrieving temporal segments within a video in constrained settings.
For example, \cite{tellex2009towards} considers retrieving video clips from a home surveillance camera using text queries which include a fixed set of spatial prepositions (``across'' and ``through'') whereas \cite{lin2014visual} considers retrieving temporal segments in 21 videos from a dashboard car camera.
In a similar vein, \cite{alayrac2016unsupervised,bojanowski2015weakly,sener2015unsupervised} consider aligning textual instructions to videos.
However, methods aligning instructions to videos are restricted to structured videos as they constrain alignment by instruction ordering.
In contrast, we consider localizing moments in an unconstrained open-world dataset with a wide array of visual concepts. 
To effectively train a moment localization model, we collect \datasetname{} which is unique because it consists of paired video moments and referring expressions.

\myparagraph{Video Summarization.}
Video summarization algorithms isolate temporal segments in a video which include important/interesting content.
Though most summarization algorithms do not include textual input (\cite{boiman2007irregularities,gygli2015video,gygli2016video2gif,yang2015unsupervised,yaohighlight}), some use text in the form of video titles \cite{liu2015multi,song2015tvsum} or user queries in the form of category labels to guide content selection \cite{sharghi2016query}.
\cite{yeung2014videoset} collects textual descriptions for temporal video chunks as a means to evaluate summarization algorithms.
However, these datasets do not include referring expressions and are limited in scope which makes them unsuitable for learning moment retrieval in an open-world setting.

\myparagraph{Video Description and Question Answering (QA).}
Video description models learn to generate textual descriptions of videos given video-description pairs.
Contemporary models integrate deep video representations with recurrent language models \cite{pan2015jointly,rohrbach2015long,venugopalan15iccv,venugopalan:naacl15,yu2015video}.
Additionally, \cite{tapaswi2015movieqa} proposed a video QA dataset which includes question/answer pairs aligned to video shots, plot synopsis, and subtitles.

YouTube and movies are popular sources for joint video-language datasets.
Video description datasets collected from YouTube include descriptions for short clips of longer YouTube videos~\cite{chen2011collecting,xu2016msr-vtt}.
Other video description datasets include descriptions of short clips sourced from full length movies~\cite{rohrbach2015dataset,torabi2015using}.
However, though YouTube clips and movie shots are sourced from longer videos, they are not appropriate for localizing distinct moments in video for two reasons.
First, descriptions about selected shots and clips are not guaranteed to be referring.
For example, a short YouTube video clip might include a person talking and the description like ``A woman is talking''.
However, the entire video could consist of a woman talking and thus the description does not uniquely refer to the clip.
Second, many YouTube videos and movies are edited, which means ``boring'' content which may be important to understand for applications like retrieving video segments from personal videos might not be present.
 
\myparagraph{Natural Language Object Retrieval.}
Natural language object retrieval  \cite{hu2015natural,mao2015generation} can be seen as an analogous task to ours, where natural language phrases are localized spatially in images, rather than temporally in videos. 
Despite similarities to natural language object retrieval, localizing video moments presents unique challenges.
For example, it often requires comprehension of temporal indicators such as ``first'' as well as a better understanding of activities.
Datasets for natural language object retrieval include \emph{referring} expressions which can uniquely localize a specific location in a image.
Descriptions in \datasetname{} uniquely localize distinct moments and are thus also referring expressions.

\myparagraph{Language Grounding in Images and Videos.} \cite{plummer2015flickr30k,rohrbach2015grounding,socher2014grounded} tackle the task of object grounding in which sentence fragments in a description are localized to specific image regions.
Work on language grounding in video is much more limited.
Language grounding in video has focused on spatially grounding objects and actions in a video \cite{lin2014visual,yu2013grounded}, or aligning textual phrases to temporal video segments \cite{regneri2013grounding,tellex2009towards}.
However prior methods in both these areas (\cite{tellex2009towards,yu2013grounded}) severely constrain natural language vocabulary (e.g., \cite{yu2013grounded} only considers four objects and four verbs) and consider constrained visual domains in small datasets (e.g., 127 videos from a fixed laboratory kitchen \cite{regneri2013grounding} and \cite{lin2014visual} only includes 520 sentences).
In contrast, \datasetname{} offers a unique opportunity to study temporal language grounding in an open-world setting with a diverse set of objects, activities, and attributes.

%% file: method.tex
\section{Moment Context Network}
\begin{figure}[]
\begin{center}
  \includegraphics[width=\linewidth]{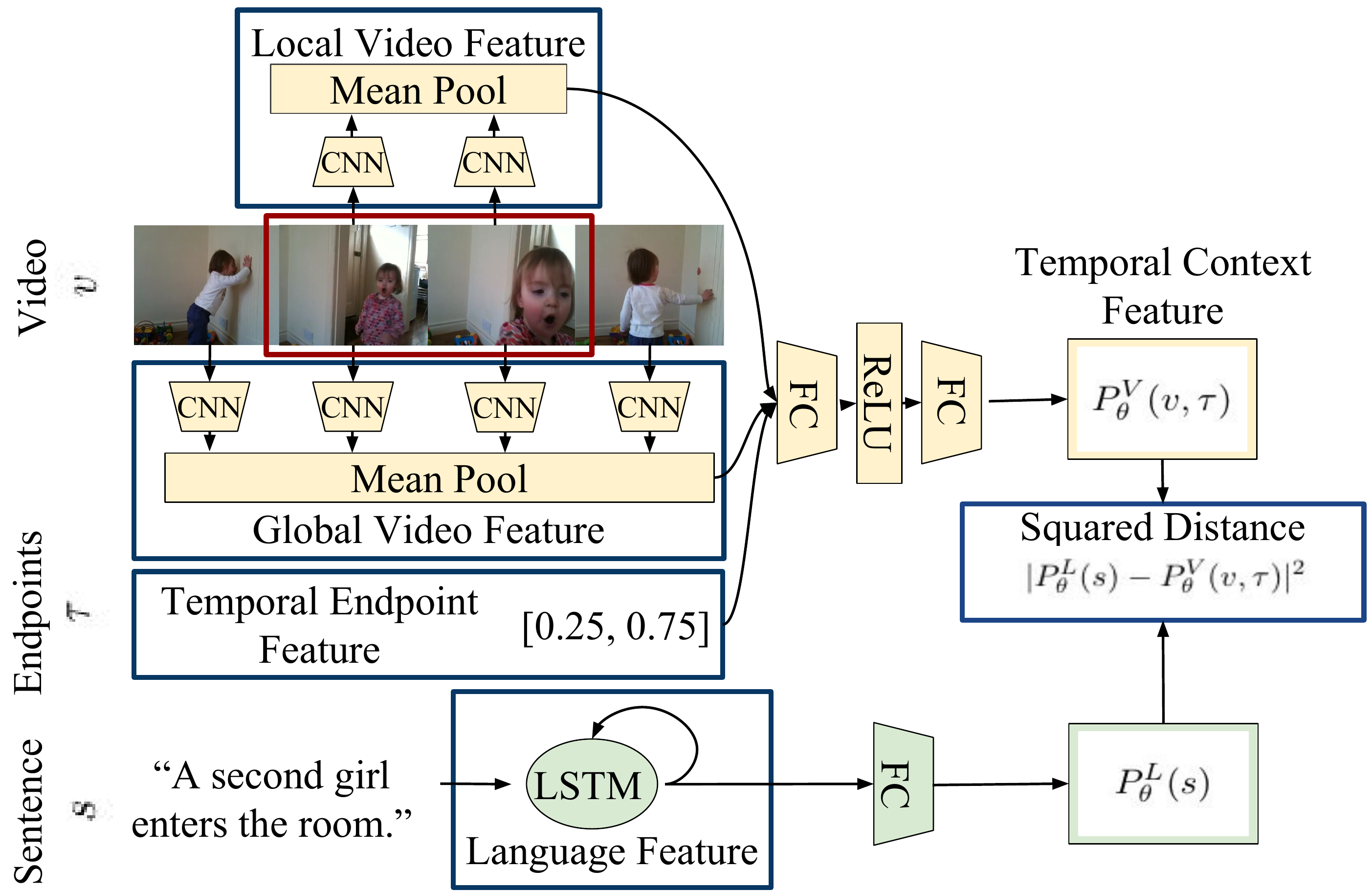}
 \end{center}
\vspace{-0.1in}
\caption{Our Moment Context Network (\modelname{}) learns a shared embedding for video temporal context features and LSTM language features.  Our  video temporal context features integrate local video features, which reflect what occurs during a specific moment, global features, which provide context for the specific moment, and temporal endpoint features which indicate when a moment occurs in a video.  We consider both appearance and optical flow input modalities, but for simplicity only show the appearance input modality here.}
\label{fig:model}
\vspace{-0.2in}
\end{figure}

Our moment retrieval model effectively localizes natural language queries in longer videos.
Given input video frames $v=\{v_t\}$,  where $t\in\{0,\dots,T-1\}$ indexes time, and a proposed temporal interval, $\hat{\tau}=\tau_{start}:\tau_{end}$, we extract visual temporal context features which encode the video moment by integrating both local features and global video context.
Given a sentence $s$ we extract language features using an LSTM~\cite{hochreiter1997long} network.
At test time our model optimizes the following objective
\begin{equation}
\hat{\tau} = \argmin_\tau D_\theta(s, v, \tau),
\end{equation}
where $D_\theta$ is a joint model over the sentence $s$, video $v$, and temporal interval $\tau$ given model parameters $\theta$ (Figure~\ref{fig:model}).

\myparagraph{Visual Temporal Context Features.}
We encode video moments into visual temporal context features by integrating local video features, which reflect what occurs within a specific moment, global video features, which provide context for a video moment, and temporal endpoint features, which indicate when a moment occurs within a longer video.
To construct local and global video features, we first extract high level video features using a deep convolutional network for each video frame, then average pool video features across a specific time span (similar to features employed by \cite{venugopalan:naacl15} for video description and \cite{torabi2016learning} for whole video retrieval).
Local features are constructed by pooling features within a specific moment and global features are constructed by averaging over all frames in a video.

When a moment occurs in a video can indicate whether or not a moment matches a specific query.
To illustrate, consider the query ``the bikers start the race.''  
We expect moments closer to the beginning of a video in which bikers are racing to be more similar to the description than moments at the end of the video in which bikers are racing.  
To encode this temporal information, we include temporal endpoint features which indicate the start and endpoint of a candidate moment (normalized to the interval $[0, 1]$).
We note that our global video features and temporal endpoint features are analogous to global image features and spatial context features frequently used in natural language object retrieval \cite{hu2015natural,mao2015generation}.

Localizing video moments often requires localizing specific activities (like ``jump'' or ``run'').
Therefore, we explore two sources of visual input modalities; appearance or RGB frames (${v_t}$) and optical flow frames (${f_t}$).
We extract $fc_7$ features from RGB frames using VGG \cite{simonyan2014very} pre-trained on ImageNet~\cite{russakovsky2015imagenet}.
We expect these features to accurately identify specific objects and attributes in video frames.
Likewise, we extract optical flow features from the penultimate layer from a competitive activity recognition model~\cite{wang2016temporal}.
We expect these features to help localize moments which require understanding action.

Temporal context features are extracted by inputting local video features, global video features, and temporal endpoint features into a two layer neural network with ReLU nonlinearities (Figure~\ref{fig:model} top).
Separate weights are learned when extracting temporal context features for RGB frames (denoted as $P_{\theta}^V$) and optical flow frames (denoted as $P_{\theta}^F$).

\myparagraph{Language Features.}  To capture language structure, we extract language features using a recurrent network (specifically an LSTM~\cite{hochreiter1997long}).
After encoding a sentence with an LSTM, we pass the last hidden state of the LSTM through a single fully-connected layer to yield embedded feature $P_\theta^L$.
Though our dataset contains over 40,000 sentences, it is still small in comparison to datasets used for natural language object retrieval (e.g., \cite{kazemzadeh2014referitgame,mao2015generation}).
Therefore, we find that representing words with dense word embeddings (specifically Glove \cite{pennington2014glove}) as opposed to one-hot encodings yields superior results when training our LSTM.

\myparagraph{Joint Video and Language Model.}
Our joint model is the sum of squared distances between embedded appearance, flow, and language features
\begin{equation}
D_\theta(s, v, \tau) = |P^V_\theta(v,\tau) - P^L_\theta(s)|^2 + 
 \eta |P^F_\theta(f,\tau)-P^L_\theta(s)|^2,
\end{equation}
where $\eta$ is a tunable (via cross validation) ``late fusion'' scalar parameter.  $\eta$ was set to $2.33$ via ablation studies.

\myparagraph{Ranking Loss for Moment Retrieval.} 
We train our model with a ranking loss which encourages referring expressions to be closer to corresponding moments than negative moments in a shared embedding space.
Negative moments used during training can either come from different segments within the same video (intra-video negative moments) or from different videos (inter-video negative moments).  Revisiting the video depicted in Figure~\ref{fig:ConceptFigure}, given a phrase ``the little girl jumps back up after falling'' many intra-video negative moments include concepts mentioned in the phrase such as ``little girl'' or ``jumps''.
Consequently, our model must learn to distinguish between subtle differences within a video.
By comparing the positive moment to the intra-video negative moments, our model can learn that localizing the moment corresponding to ``the little girl jumps back up after falling'' requires more than just recognizing an object (the girl) or an action (jumps). For training example $i$ with endpoints $\tau_i$, we define the following intra-video ranking loss
\begin{equation}
\mathcal{L}_i^{intra}(\theta)=\sum_{n\in\Gamma\setminus\tau^{i}} \mathcal{L}^R\left(D_\theta(s^{i}, v^{i},\tau^{i}), D_\theta(s^{i}, v^{i},n)\right),
\end{equation}
where $\mathcal{L}^R(x,y)=\max(0,x-y+b)$ is the ranking loss, $\Gamma$ are all possible temporal video intervals, and $b$ is a margin. 
Intuitively, this loss encourages text queries to be closer to a corresponding video moment than all other possible moments from the same video.

Only comparing moments within a single video means the model must learn to differentiate between subtle differences without learning how to differentiate between broader semantic concepts (e.g., ``girl'' vs.\ ``sofa'').
Hence, we also compare positive moments to inter-video negative moments which generally include substantially different semantic content.  
When selecting inter-video negative moments, we choose negative moments which have the same start and end points as positive moments.  
This encourages the model to differentiate between moments based on semantic content, as opposed to when the moment occurs in the video.
During training we do not verify that inter-video negatives are indeed true negatives.
However, the language in our dataset is diverse enough that, in practice, we observe that randomly sampled inter-video negatives are generally true negatives. 
For training example $i$, we define the following inter-video ranking loss
\begin{equation}
\mathcal{L}_i^{inter}(\theta)=\sum_{j\neq i} \mathcal{L}^R\left(D_\theta(s^{i}, v^{i},\tau^{i}), D_\theta(s^{i}, v^{j},\tau^{i})\right).
\end{equation}
This loss encourages text queries to be closer to corresponding video moments than moments outside the video, and should thus learn to differentiate between broad semantic concepts.
Our final inter-intra video ranking loss is
\begin{align}
\mathcal{L}(\theta) = \lambda \sum_i\mathcal{L}_i^{intra}(\theta) + (1 - \lambda)\sum_i\mathcal{L}_i^{inter}(\theta),
\end{align}
where $\lambda$ is a weighting parameter chosen through cross-validation.

%% file: dataset.tex
\section{The \datasetname{} Dataset}
\label{sec:dataset}

\begin{figure*}[t]
\begin{center}
  \includegraphics[width=\linewidth]{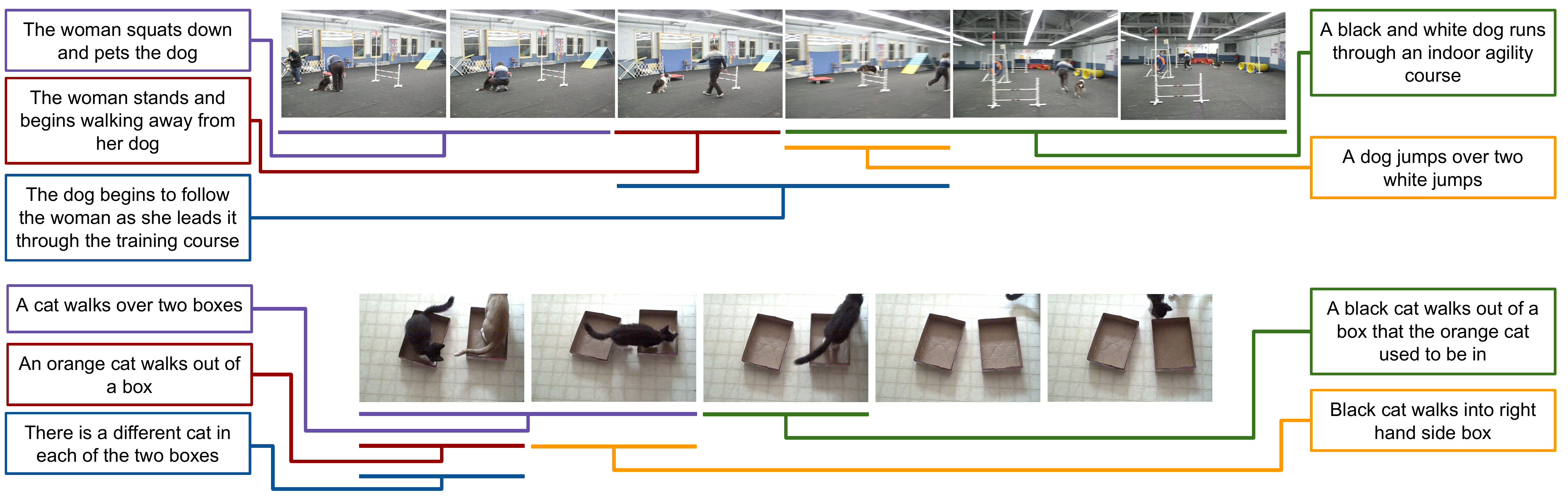}
 \end{center}
\vspace{-0.1in}
\caption{Example videos and annotations from our Distinct Describable Moments (DiDeMo) dataset.  Annotators describe moments with varied language (e.g., ``A cat walks over two boxes'' and ``An orange cat walks out of a box'').
Videos with multiple events (top) have annotations which span all five-second segments.
Other videos have segments in which no distinct event takes place (e.g., the end of the bottom video in which no cats are moving).
}
\label{fig:exampleAnnotations}
\vspace{-0.2in}
\end{figure*}

A major challenge when designing algorithms to localize moments with natural language is that there is a dearth of large-scale datasets which consist of referring expressions and localized video moemnts.
To mitigate this issue, we introduce the Distinct Describable Moments (\datasetname{}) dataset which includes over 10,000 25-30 second long personal videos with over 40,000 localized text descriptions.
Example annotations are shown in Figure~\ref{fig:exampleAnnotations}.

\subsection{Dataset Collection}

To ensure that each description is paired with a single distinct moment, we collect our dataset in two phases (similar to how \cite{kazemzadeh2014referitgame} collected text to localize image regions).
First, we asked annotators to watch a video, select a moment, and describe the moment such that another user would select the same moment based on the description.
Then, descriptions collected in the first phase are validated by asking annotators to watch videos and mark moments that correspond to collected descriptions.

\myparagraph{Harvesting Personal Videos.}
We randomly select over 14,000 videos from YFCC100M \cite{thomee2015new} which contains over 100,000 Flickr videos with a Creative Commons License.
To ensure harvested videos are unedited, we run each video through a shot detector based on the difference of color histograms in adjacent frames \cite{mas2003video} then manually filter videos which are not caught.
Videos in \datasetname{} represent a diverse set of real-world videos, which include interesting, distinct moments, as well as uneventful segments which might be excluded from edited videos.

\myparagraph{Video Interface.}
Localizing text annotations in video is difficult because the task can be ambiguous and users must digest a 25-30s video before scrubbing through the video to mark start and end points.
To illustrate the inherent ambiguity of our task, consider the phrase ``The woman leaves the room.''
Some annotators may believe this moment begins as soon as the woman turns towards the exit, whereas others may believe the moment starts as the woman's foot first crosses the door threshold.
Both annotations are valid, but result in large discrepancies between start and end points.

To make our task less ambiguous and speed up annotation, we develop a user interface in which videos are presented as a timeline of temporal segments.
Each segment is displayed as a gif, which plays at 2x speed when the mouse is hovered over it.
Following \cite{yeung2014videoset}, who collected localized text annotations for summarization datasets, we segment our videos into 5-second segments.
Users select a moment by clicking on all segments which contain the moment.
To validate our interface, we ask five users to localize moments in ten videos using our tool and a traditional video scrubbing tool.
Annotations with our gif-based tool are faster to collect (25.66s vs.\ 38.48s).
Additionally, start and end points marked using the two different tools are similar.
The standard deviation for start and end points marked when using the video scrubbing tool (2.49s) is larger than the average difference in start and end points marked using the two different tools (2.45s). 

\begin{table*}[t]
\setlength{\tabcolsep}{0.5pt} %
\setlength\extrarowheight{1.8pt}
\centering
\scriptsize
\footnotesize
\rowcolors{2}{white}{blue!10}
\resizebox{\linewidth}{!}{%

  \begin{tabular}{LMMMMMMM}
  \toprule
    Dataset  &\# Videos/\# Clips & \# Sentences & Video Source & \ \ Domain \ \ & Temporal Localization &  Un-Edited & Referring Expressions\\
    \hline
    YouCook  \cite{das2013thousand} & 88/- & 2,668 & YouTube & Cooking & & & \\
    Charades \cite{sigurdsson2016hollywood} & 10,000/- & 16,129 & Homes &
                     \scriptsize{Daily activities} &  & \checkmark&\\
    TGIF \cite{li2016tgif} & 100,000 /-& 125,781 & Tumblr GIFs & Open & & &\\
    MSVD \cite{chen2011collecting} & 1,970/1,970 & 70,028 & YouTube  &
                     Open& \checkmark  & & \\
    MSR-VTT \cite{xu2016msr-vtt}& 7,180/10,000 & 200,000 & YouTube  &
                     Open& \checkmark&  &\\
    LSMDC 16 \cite{lsmdc}& 200/128,085  & 128,085 & Movie  &
                        Open& \checkmark &   &\\
     TV Dataset \cite{yeung2014videoset} & 4/1,034 & 1,034  &  TV Shows  & TV Shows & \checkmark &  &  \\

    KITTI \cite{lin2014visual} & 21/520  & 520 & Car Camera &   Driving &\checkmark & \checkmark & \\
    TACoS \cite{regneri2013grounding,rohrbach2013translating} & 123/7,206 & 18,227 & Lab Kitchen &
Cooking & \checkmark & \checkmark &\\
    TACoS multi-level\cite{rohrbach2014coherent} & 185/14.105 & 52,593 & Lab Kitchen & Cooking & \checkmark & \checkmark & \\
    UT Egocentric \cite{yeung2014videoset} & 4/11,216 & 11,216 &  Egocentric & Daily Activities &\checkmark  & \checkmark &  \\
    Disneyland \cite{yeung2014videoset} & 8/14,926 & 14,916 & Egocentric & Disneyland &\checkmark  & \checkmark & \\
    \datasetname{} & 10,464/26,892 &  40,543 &  Flickr & Open & \checkmark & \checkmark & \checkmark \\
\bottomrule

  \end{tabular}
  }
  \vspace{-.1cm}

  \caption{Comparison of \datasetname{} to other video-language datasets.  \datasetname{} is unique because it includes a validation step ensuring that descriptions are referring expressions. }
\label{tab:datasetComparison}
  \vspace{-.1cm}
\end{table*}

\myparagraph{Moment Validation.}
After annotators describe a moment, we ask three additional annotators to localize the moment given the text annotation and the same video.
To accept a moment description,  we require that at least three out of four annotators (one describer and three validators) be in agreement.
We consider two annotators to agree if one of the start \emph{or} end point differs by at most one gif.

\subsection{\datasetname{} Summary}

Table~\ref{tab:datasetComparison} compares our Distinct Describable Moments (\datasetname{}) dataset to other video-language datasets.
Though some datasets include temporal localization of natural language, these datasets do not include a verification step to ensure that descriptions refer to a single moment.
In contrast, our verification step ensuring that descriptions in \datasetname{} are \emph{referring expressions}, meaning that they refer to a specific moment in a video.

\myparagraph{Vocabulary.}
Because videos are curated from Flickr, \datasetname{} reflects the type of content people are interested in recording and sharing.
Consequently, \datasetname{} is human-centric with words like ``baby'', ``woman'', and ``man'' appearing frequently.
Since videos are randomly sampled, \datasetname{} has a long tail with words like ``parachute'' and ``violin'', appearing infrequently (28 and 38 times). %

Important, distinct moments in a video often coincide with specific camera movements.
For example, ``the camera pans to a group of friends'' or ``zooms in on the baby'' can describe distinct moments.
Many moments in personal videos are easiest to describe in reference to the viewer (e.g., ``the little boy runs towards the camera'').
In contrast to other dataset collection efforts \cite{chen2011collecting}, we allow annotations to reference the camera, and believe such annotations may be helpful for applications like text-assisted video editing.

Table \ref{tab:wordTypeComp} contrasts the kinds of words used in \datasetname{} to two natural language object retrieval datasets \cite{kazemzadeh2014referitgame,mao2015generation} and two video description datasets \cite{lsmdc,xu2016msr-vtt}.
The three left columns report the percentage of sentences which include camera words (e.g., ``zoom'', ``pan'', ``cameraman''), temporal indicators (e.g., ``after'' and ``first''), and spatial indicators (e.g., ``left'' and ``bottom'').
We also compare how many words belong to certain parts of speech (verb, noun, and adjective) using the natural language toolkit part-of-speech tagger \cite{bird2009natural}.
\datasetname{} contains more sentences with temporal indicators than natural language object retrieval and video description datasets, as well as a large number of spatial indicators.
\datasetname{} has a higher percentage of verbs than natural language object retrieval datasets, suggesting understanding action is important for moment localization in video.

\myparagraph{Annotated Time Points.}
Annotated segments can be any contiguous set of gifs.
Annotators generally describe short moments with 72.34\% of descriptions corresponding to a single gif and 22.26\% corresponding to two contiguous gifs.
More annotated moments occur at the beginning of a video than the end.
This is unsurprising as people generally choose to begin filming a video when something interesting is about to happen.
In 86\% of videos annotators described multiple distinct moments with an average of 2.57 distinct moments per video.

\begin{table}[t]
\setlength{\tabcolsep}{5pt} %
\setlength\extrarowheight{1.5pt}
\centering
\small
\footnotesize
\scriptsize
\rowcolors{3}{white}{blue!10}
\resizebox{\linewidth}{!}{%
  \label{tab:tableAblationResults}
  \begin{tabular}{lccc|ccc}
    \toprule
     &  \multicolumn{3}{c}{\% Sentences} & \multicolumn{3}{c}{\% Words} \\
      & Camera & Temp. & Spatial & Verbs & Nouns & Adj.\\
     \midrule
    ReferIt \cite{kazemzadeh2014referitgame} & 0.33 & 1.64 & 43.13 & 5.88 & 52.38 & 11.54\\
    RefExp \cite{mao2015generation} & 1.88 & 1.00 & 15.11 & 8.97 &36.26 & 11.82\\
    MSR-VTT \cite{xu2016msr-vtt} & 2.10 & 2.03 & 1.24 &18.77 &36.95 & 5.12 \\   
    LSMDC 16 \cite{lsmdc} & 1.09 & 7.58 & 1.49 & 13.71 &37.44 & 3.99 \\   
    \datasetname{} & 19.69 & 18.42 & 11.62 & 16.06 & 35.26 & 7.89\\
    \bottomrule
  \end{tabular}
  }
    \normalsize
  \caption{\datasetname{} contains more camera and temporal words than natural language object recognition datasets \cite{kazemzadeh2014referitgame,mao2015generation} or video description datasets \cite{xu2016msr-vtt,lsmdc}.  Additionally, verbs are more common in \datasetname{} than in natural language object retrieval datasets suggesting natural language moment retrieval relies more heavily on recognizing actions than natural language object retrieval.}
  \label{tab:wordTypeComp}
  \vspace{-.3cm}
\end{table}

%% file: results.tex
\section{Evaluation}

In this section we report qualitative and quantitative results on \datasetname{}.  First, we describe our evaluation criteria and then evaluate against baseline methods.

\begin{table}[t]
\setlength{\tabcolsep}{5pt} %
\setlength\extrarowheight{1.8pt}
\centering
\small
\footnotesize
\rowcolors{3}{white}{blue!10}
\resizebox{\linewidth}{!}{%

  \begin{tabular}{llrrr}
    \toprule
     \multicolumn{5}{c}{\textbf{Baseline Comparison (Test Set)}}\\
     \midrule
     & Model & Rank$@$1 & Rank$@$5 & mIoU \\
    \midrule
    \midrule
     1 & Upper Bound & 74.75 & 100.00 & 96.05 \\    
     2 & Chance &  3.75 & 22.50 & 22.64 \\
     3 & Moment Frequency Prior & 19.40 & 66.38 & 26.65 \\
     4 & CCA & 18.11 & 52.11 & 37.82 \\
     5 & Natural Lang. Obj. Retrieval \cite{hu2015natural} & 16.20  & 43.94 & 27.18 \\
     6 & Natural Lang. Obj. Retrieval \cite{hu2015natural} (re-trained) & 15.57 & 48.32 & 30.55 \\
    \midrule
    7  & \modelname{} (ours)  & \textbf{28.10} & \textbf{78.21} & \textbf{41.08} \\
    \bottomrule 
    \midrule
    \rowcolor{white}
    \multicolumn{5}{c}{\textbf{Ablations (Validation Set)}}\\
    8 & LSTM-RGB-local &  13.10 & 44.82 & 25.13\\
    9 & LSTM-Flow-local &  18.35 & 56.25 & 31.46 \\
    10 & LSTM-Fusion-local &  18.71 & 57.47 & 32.32 \\
    11 & LSTM-Fusion + global &  19.88 & 62.39 & 33.51\\
    12 & LSTM-Fusion + global + tef (\modelname{}) &  \textbf{27.57} & \textbf{79.69} & \textbf{41.70}\\
    \midrule
  
  \end{tabular}
  }
    \normalsize
  \caption{Our Moment Context Network (\modelname{}) outperforms baselines (rows 1-6) on our test set.  We show ablation studies on our validation set in rows 8-12.  Both flow and RGB modalities are important for good performance (rows 8-10).  Global video features and temporal endpoint features (tef) both lead to better performance (rows 10-12).
}
  \label{tab:results}
  \vspace{-.3cm}
\end{table}

\myparagraph{Metrics: Accounting for Human Variance.}
Our model ranks candidate moments in a video based on how well they match a text description.
Candidate moments come from the temporal segments defined by the gifs used to collect annotations.
A 30 second video will be broken into six five-second gifs.
Moments can include any contiguous set of gifs, so a 30-second video contains 21 possible moments.
We measure the performance of each model with Rank$@$1 (R$@$1), Rank$@$5 (R$@$5), and mean intersection over union (mIoU).
Instead of consolidating all human annotations into one ground truth, we compute the score for a prediction and each human annotation for a particular description/moment pair.
To account for outlier annotations, we consider the highest score among sets of annotations $A'$ where $A'$ are the four-choose-three combinations of all four annotations $A$.
Hence, our final score for a prediction $P$ and four human annotations $A$ using metric $M$ is:
	$score(P,A) = \max_{A' \in \binom{A}{3}} \frac{1}{3}\sum_{a \in A'} M(P,a)$.
As not all annotators agree on start and end points it is impossible to achieve 100\% on all metrics (c.f., upper bounds in Table~\ref{tab:results}).

\myparagraph{Baseline: Moment Frequency Prior.}
Though annotators may mark any contiguous set of gifs as a moment, they tend to select short moments toward the beginning of videos.
The moment frequency prior selects moments which correspond to gifs most frequently described by annotators.

\myparagraph{Baseline: CCA.}
Canonical correlation analysis (CCA) achieves competitive results for both natural language image~\cite{klein2015associating} and object~\cite{plummer2015flickr30k} retrieval tasks.
We use the CCA model of~\cite{klein2015associating} and employ the same visual features as the MCN model.
We extract language features from our best MCN language encoder for fair comparison.

\myparagraph{Baseline: Natural Language Object Retrieval.} 
Natural language object retrieval models localize objects in a text image.
We verify that localizing objects is not sufficient for moment retrieval by running a natural language object retrieval model \cite{hu2015natural} on videos in our test set.
For every tenth frame in a video, we score candidate bounding boxes with the object retrieval model proposed in \cite{hu2015natural} and compute the score for a frame as the maximum score of all bounding boxes.
The score for each candidate moment is the average of scores for frames within the moment.
Additionally, we re-train \cite{hu2015natural} using the same feautures used to train our MCN model; instead of candidate bounding boxes, we provide candidate temporal chunks and train with both appearance and flow input modalities.
More details, baselines, and ablations can be found in our appendix.

\myparagraph{Implementation Details.}
\datasetname{} videos are split into training (8,395), validation (1,065), and testing (1,004) sets.
Videos from a specific Flickr user only appear in one set.
All models are implemented in Caffe~\cite{jia2014caffe} and have been publicly released \footnote{\tt\small \url{https://people.eecs.berkeley.edu/~lisa_anne/didemo.html}}.
SGD (mini-batch size of 120) is used for optimization and all hyperparamters, such as embedding size (100), margin (0.1), and LSTM hidden state size (1000), are chosen through ablation studies. 
\subsection{Results}

Table~\ref{tab:results} compares different variants of our proposed retrieval model to our baselines.
Our ablations demonstrate the importance of our temporal context features and the need for both appearance and optical flow features.

\myparagraph{Baseline Comparison.} Rows 1-7 of Table~\ref{tab:results} compare the Moment Context Network (\modelname{}) model to baselines on our test set.
Though all baselines we trained (lines 4-6) have similar R$@$1 and R$@$5 performance, CCA performs substantially better on the mIoU metric.
Scoring video segments based on the scores from a natural language object retrieval model \cite{hu2015natural} does fairly well, performing similarly to the same model retrained with our features.
This suggests that pre-training with a dataset designed for natural language object retrieval and incorporating spatial localization into our model could improve results. 
We believe that retraining \cite{hu2015natural} leads to poor results on our dataset because it relies on sentence generation rather than directly retrieving a moment.
Additionally, our model does substantially better than the moment frequency prior.

\myparagraph{Visual Temporal Context Feature.} Rows 9-12 of Table~\ref{tab:results} demonstrate the importance of temporal context for moment retrieval.
The inclusion of both the global video feature and temporal endpoint feature increase performance considerably.  
Additionally, we find that combining both appearance and optical flow features is important for best performance.

\begin{figure*}[t]
\begin{center}
  \includegraphics[width=\linewidth]{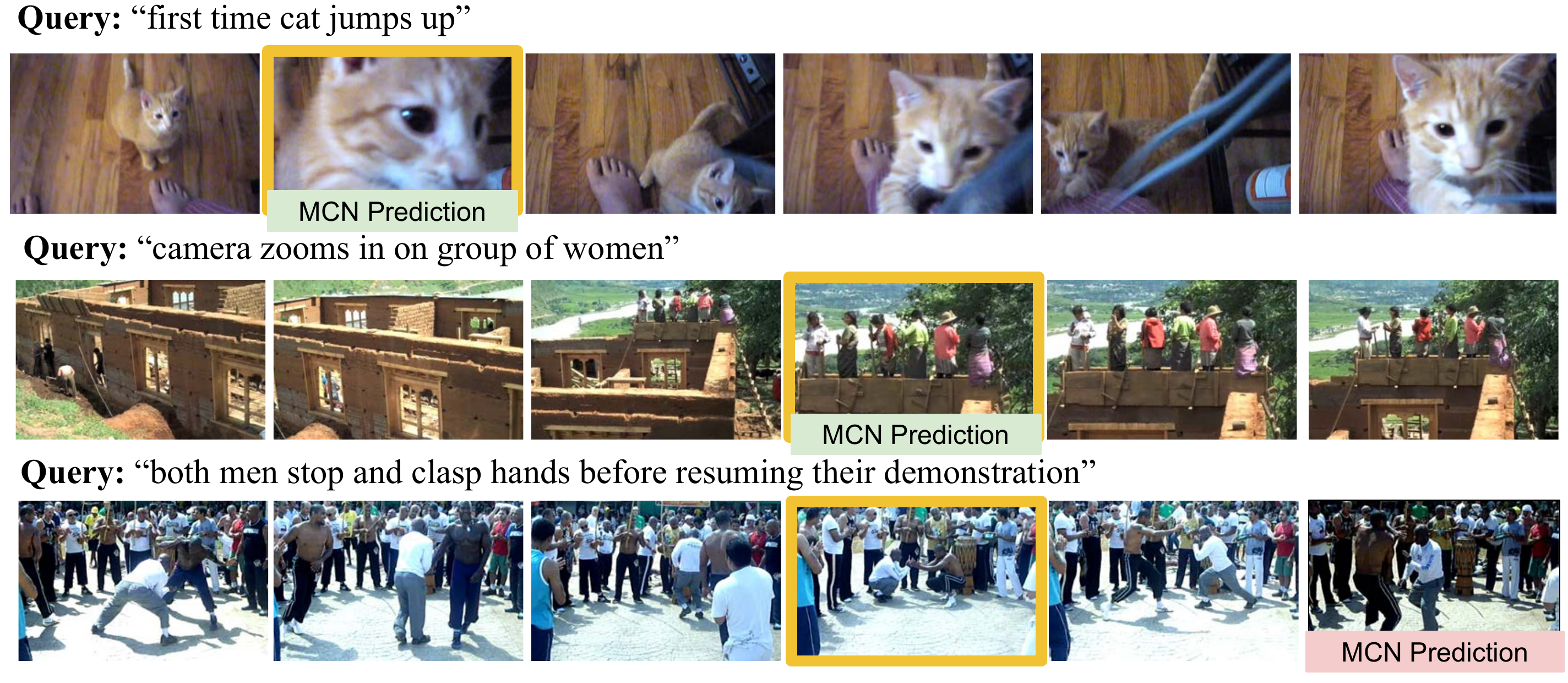}
 \end{center}
  \vspace{-.6cm}
\caption{Natural language moment retrieval results on \datasetname{}.  Ground truth moments are outlined in yellow.  The Moment Context Network (\modelname{}) localizes diverse descriptions which include temporal indicators, such as ``first'' (top), and camera words, such as ``camera zooms'' (middle).}
\label{fig:results}
  \vspace{-.3cm}
\end{figure*}

\begin{figure}[]
\begin{center}
  \includegraphics[width=\linewidth]{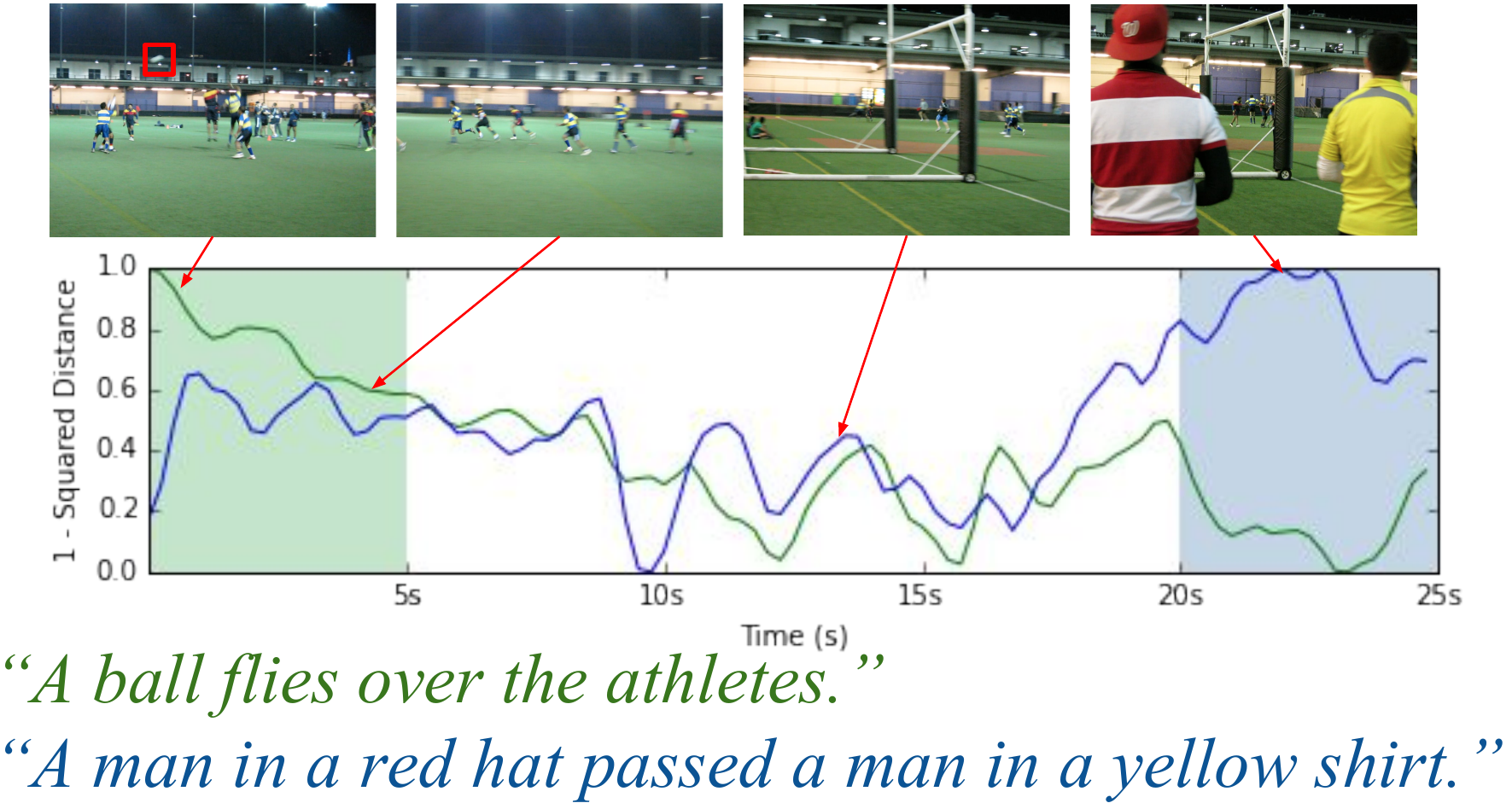}
 \end{center}
  \vspace{-.4cm}
\caption{\modelname{} correctly retrieves two different moments (light green rectangle on left and light blue rectangle on right).  Though our ground truth annotations are five-second segments, we can evaluate with more fine-grained temporal proposals at test time.  This gives a better understanding of when moments occur in video (e.g., ``A ball flies over the athletes'' occurs at the start of the first temporal segment).}
\label{fig:multiscale}
  \vspace{-.6cm}
\end{figure}

\myparagraph{Qualitative Results.} Figure~\ref{fig:results} shows moments predicted by \modelname{}.
Our model is capable of localizing a diverse set of moments including moments which require understanding temporal indicators like ``first'' (Figure~\ref{fig:results} top) as well as moments which include camera motion (Figure~\ref{fig:results} middle).  More qualitative results are in our appendix.

\myparagraph{Fine-grained Moment Localization}  Even though our ground truth moments correspond to five-second chunks, we can evaluate our model on smaller temporal segments at test time to predict moment locations with finer granularity.
Instead of extracting features for a five second segment, we evaluate on individual frames extracted at $\sim$ 3 fps. 
Figure~\ref{fig:multiscale} includes an example in which two text queries (``A ball flies over the athletes'' and ``A man in a red hat passed a man in a yellow shirt'') are correctly localized by our model.
The frames which best correspond to ``A ball flies over the athletes'' occur in the first few seconds of the video and the moment ``A man in a red hat passed a men in a yellow shirt'' finishes before the end point of the fifth segment.
More qualitative results are in our appendix.

\myparagraph{Discussion.} 
We introduce the task of localizing moments in video with natural language in a challenging, open-world setting.
Our Moment Context Network (\modelname{}) localizes video moments by harnessing local video features, global video features, and temporal endpoint features.
To train and evaluate natural language moment localization models, we collect  \datasetname{}, which consists of over 40,000 pairs of localized moments and referring expressions.
Though \modelname{} properly localizes many natural language queries in video, there are still many remaining challenges.
For example, modeling complex (temporal) sentence structure is still very challenging (e.g.,  our model fails to localize ``dog stops, then starts rolling around again''). 
Additionally, DiDeMo has a long-tail distribution with rare activities, nouns, and adjectives. 
More advanced (temporal) language reasoning and improving generalization to previously unseen vocabulary are two potential future directions.

%% file: appendix.tex
This appendix includes the following material:

\begin{enumerate}
\item Qualitative examples illustrating when global video features and tef features improve performance.
\item Qualitative examples contrasting RGB and flow input modalities.
\item Additional qualitative examples using the full Moment Context Network.  See {\tt\small \url{https://www.youtube.com/watch?v=MRO7_4ouNWU}} for a video example.
\item Additional baselines.
\item Ablation of inter-intra negative loss.
\item Results when training without a language feature. 
\item List of words used to generate numbers in Table 2 of the main paper.
\item Qualitative video retrieval experiment.  See {\tt\small \url{https://www.youtube.com/watch?v=fuz-UBvgapk}} for a video example.
\item Discussion on ambiguity of annotations and our metrics.
\item Histrogram showing the moments annotators mark in our dataset.
\item Example video showing our annotation tool (see {\tt\small \url{https://www.youtube.com/watch?v=vAvT5Amp408}} and {\tt\small \url{https://www.youtube.com/watch?v=9WWgndeEjMU}}.
\end{enumerate}

\section{Impact of Global Video Features and TEF Features}
In the main paper we quantitatively show that global video features and tef features improve model performance.
Here, we highlight qualitative examples where the global video features and tef features lead to better localization.

\begin{figure*}[t]
\begin{center}
  \includegraphics[width=\linewidth]{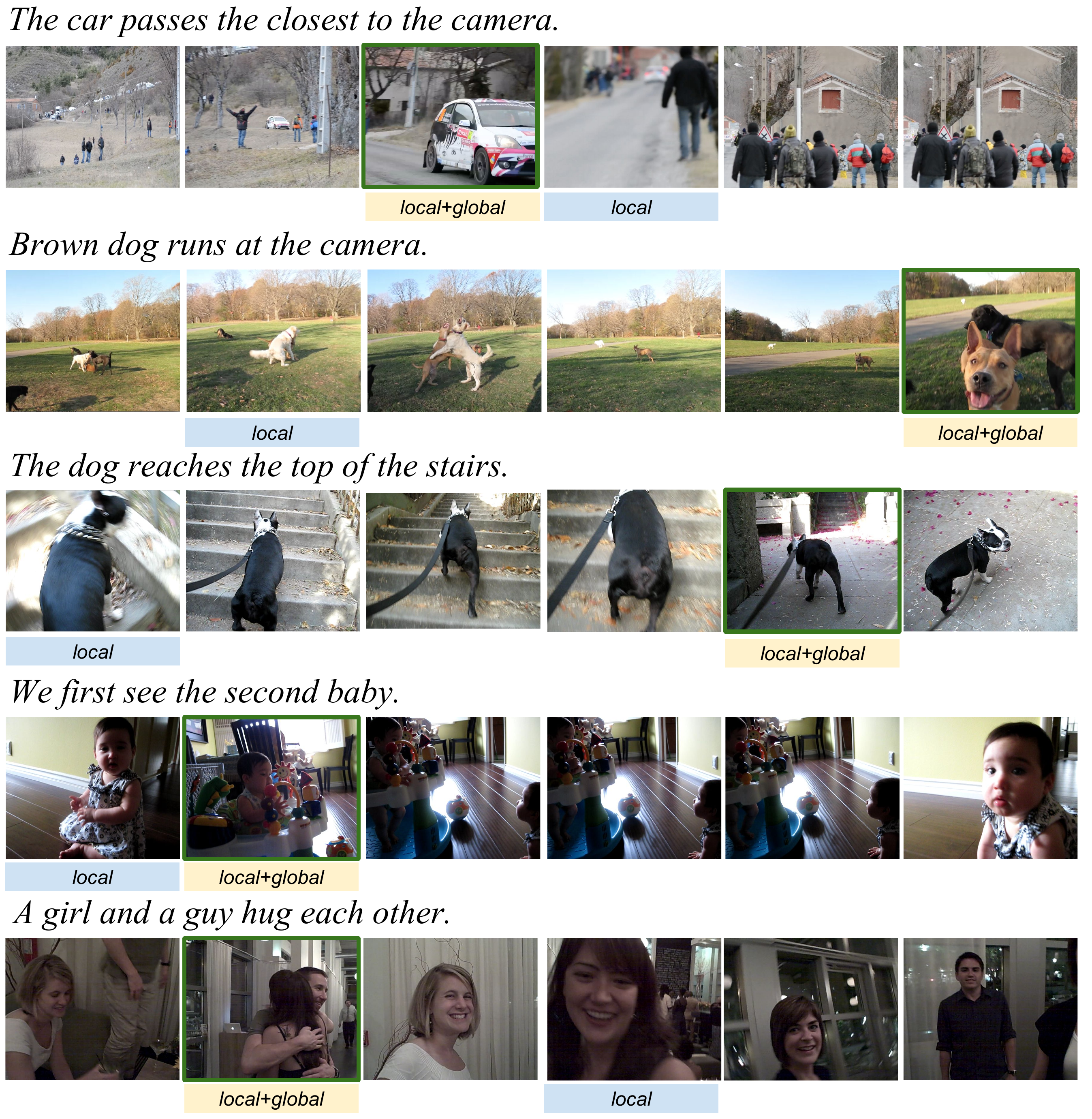}
 \end{center}
\caption{Comparison of moments which are correctly retrieved when including global context, but not when only using local video features.  Ground truth moments are outlined in green.
 Global video features improve results for a variety of moments.  For moments like ``the car passes the closest to the camera'', it is not enough to identify a car but to understand when the car is closer to the camera than in any other moment.  For moments like ``brown dog runs at the camera'', the model must not only identify when the brown dog is running, but when it runs towards the camera.}
\label{fig:global_v_local}
\end{figure*}

Figure~\ref{fig:global_v_local} shows examples in which including global context improves performance.
Examples like ``The car passes the closest to the camera'' require context to identify the correct moment.
This is sensible as the word ``closest'' is comparative in nature and determining when the car is closest requires viewing the entire video.
Other moments which are correctly localized with context include ``we first see the second baby'' and ``the dog reaches the top of the stairs''.

\begin{figure*}[t]
\begin{center}
  \includegraphics[width=\linewidth]{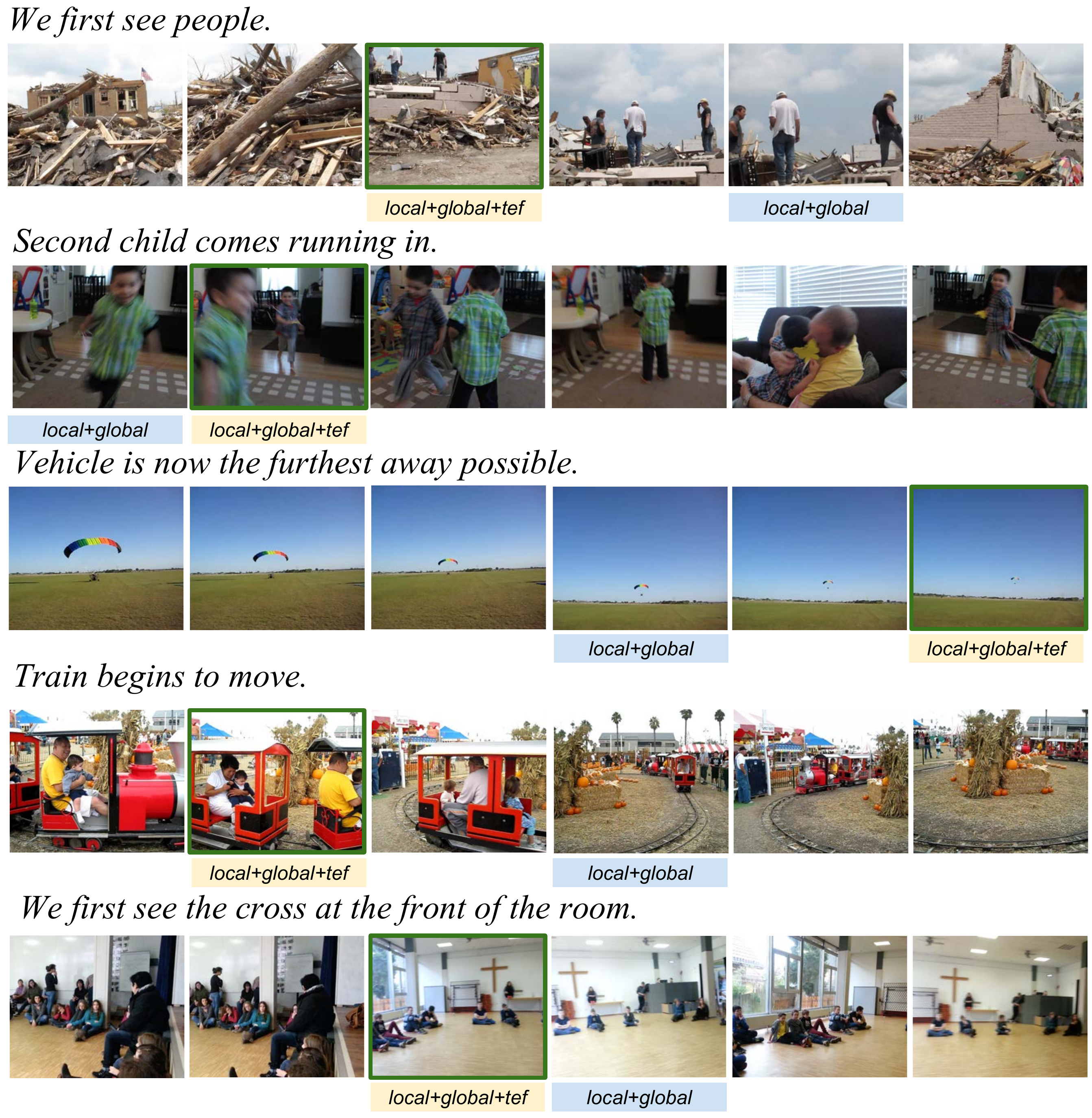}
 \end{center}
\caption{Comparison of moments which are correctly retrieved when including the temporal endpoint feature (tef), but not when only using local and global video features.  Ground truth moments are outlined in green.  For moments like ``we first see the people'' the model without tef retrieves a video moment with people, but fails to retrieve the moment when the people first appear.  Likewise, for moments like ``train begins to move'', the model without tef retrieves a video moment in which the train is moving, but not a moment in which the train begins to move.}
\label{fig:global_v_tef}
\end{figure*}

Figure~\ref{fig:global_v_tef} shows examples in which including temporal endpoint features (tef) correctly localizes a video moment.
For moments like ``we first see the people'' the model without tef retrieves a video moment with people, but fails to retrieve the moment when the people first appear.  
Without the tef, the model has no indication of \textit{when} a moment occurs in a video.  
Thus, though the model can identify if there are people in a moment, the model is unable to determine when the people first appear.
Likewise, for moments like ``train begins to move'', the model without tef retrieves a video moment in which the train is moving, but not a moment in which the train begins to move.

\section{RGB and Flow Input Modalities}

\begin{figure*}[t]
\begin{center}
  \includegraphics[width=\linewidth]{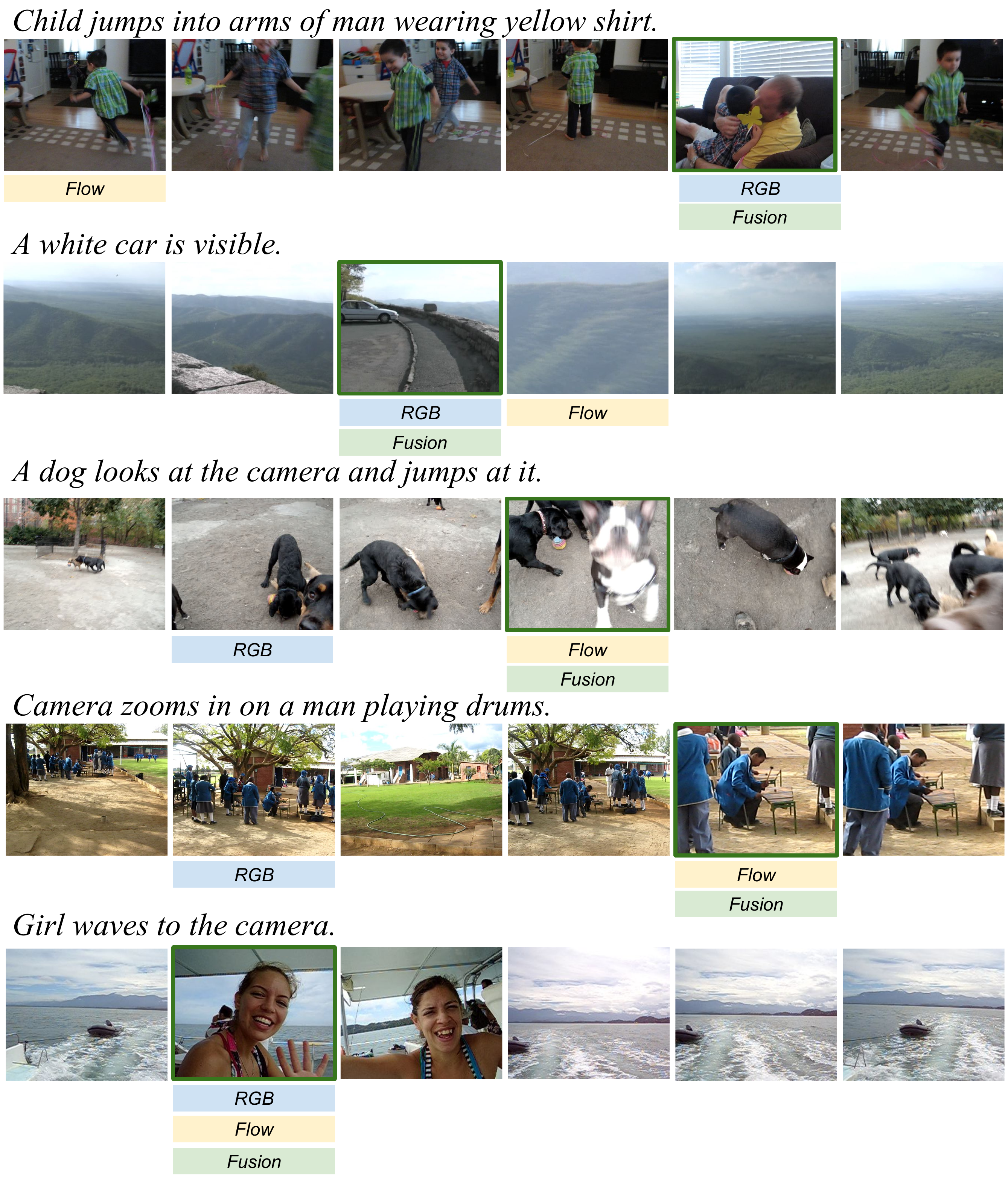}
 \end{center}
\caption{Comparison of moments retrieved using different input modalities (ground truth marked in green).  For queries like ``A white car is visible'' which require recognizing an object, a network trained with RGB performs better whereas for queries like ``Camera zooms in on a man playing drums'' which require understanding movement, a network trained with optical performs better.  For some queries, networks trained with either RGB or optical flow retrieve the correct moment.}
\label{fig:flow_rgb_fusion}
\end{figure*}

In the main paper, we demonstrate that RGB and optical flow inputs are complementary. 
Here we show a few examples which illustrate how RGB and flow input modalities complement each other.
Figure~\ref{fig:flow_rgb_fusion} compares a model trained with RGB input and a model trained with optical flow input (both trained with global video features and tef).
We expect the model trained with RGB to accurately localize moments which require understanding the appearance of objects and people in a scene, such as ``child jumps into arms of man wearing yellow shirt'' (Figure~\ref{fig:flow_rgb_fusion} top row). 
We expect the model trained with flow to better localize moments which require understanding of motion (including camera motion) such as ``a dog looks at the camera and jumps at it'' and ``camera zooms in on a man playing the drums'' (Figure~\ref{fig:flow_rgb_fusion} row 3 and 4).
Frequently, both RGB and optical flow networks can correctly localize a moment (Figure~\ref{fig:flow_rgb_fusion} bottom row).
However, for best results we take advantage of the complimentary nature of RGB and optical flow input modalities in our fusion model.

\section{Qualitative Results for MCN}

\begin{figure*}[t]
\begin{center}
  \includegraphics[width=\linewidth]{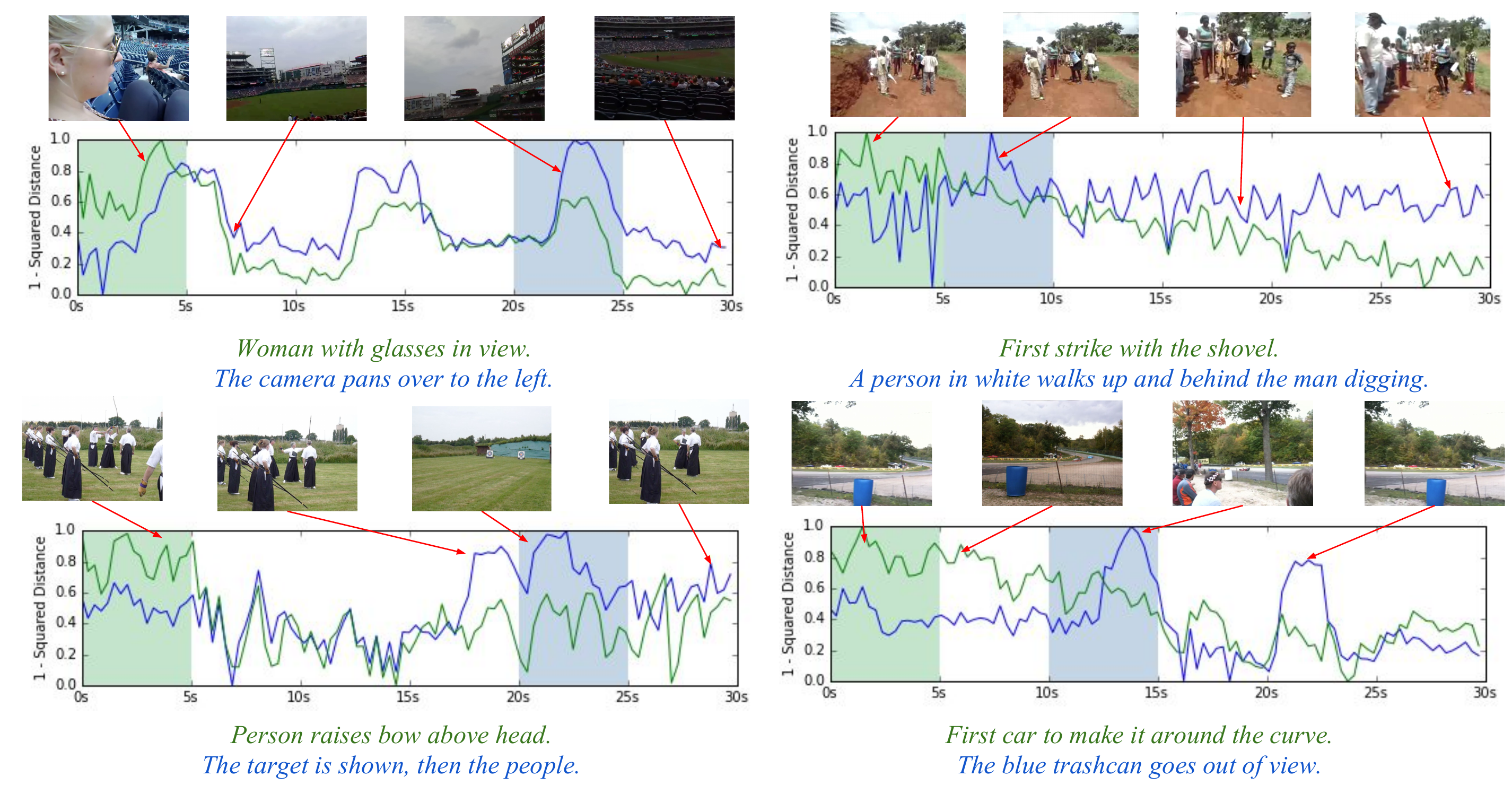}
 \end{center}
\caption{Comparison of similarity between text queries and finegrained temporal segments.  Though ground truth annotations correspond to five second segments, evaluation with more finegrained segments at test time can provide better insight about where a moment occurs within a specific segment and also provide insight into which other parts of a video are similar to a given text query.}
\label{fig:finegrained}
\end{figure*}

Figure~\ref{fig:finegrained} shows four videos in which we evaluate with fine-grained temporal windows at test time.
Observing the plots in Figure~\ref{fig:finegrained} provides insight into the exact point at which a moment occurs.
For example, our model correctly localizes the phrase ``the blue trashcan goes out of view'' (Figure~\ref{fig:finegrained} bottom right).
The finegrained temporal segments that align best with this phrase occur towards the end of the third segment (approximately 14s).
Furthermore, Figure~\ref{fig:finegrained} provides insight into which parts of the video are most similar to the text query, and which parts are most dissimilar.
For example, for the phrase ``the blue trashcan goes out of view'', there are two peaks; the higher peak occurs when the blue trashcan goes out of view, and the other peak occurs when the blue trashcan comes back into view.

\begin{figure*}[t]
\begin{center}
  \includegraphics[width=\linewidth]{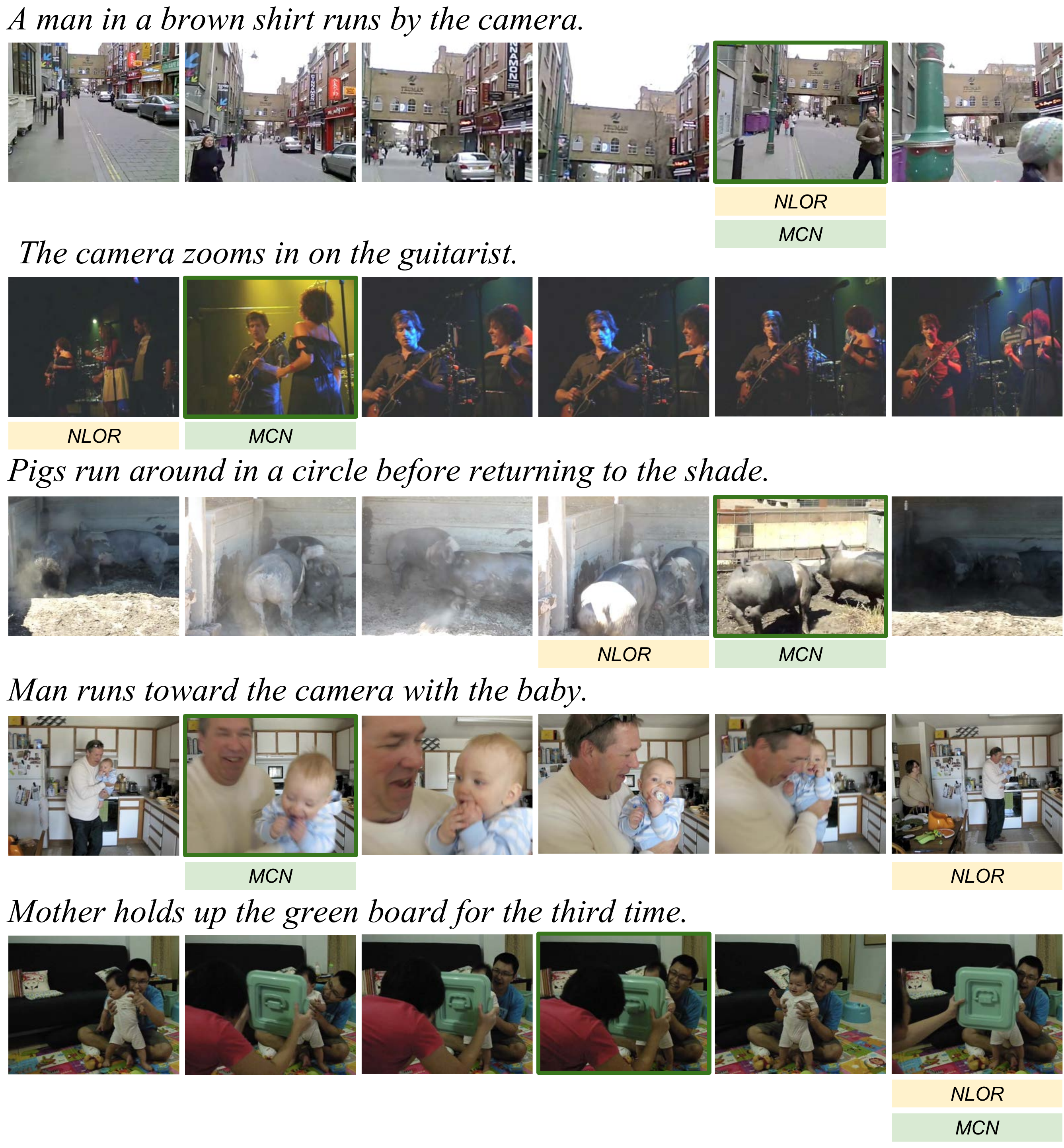}
 \end{center}
\caption{We compare our Moment Context Network (MCN) model to a model trained for natural language object retrieval (NLOR).  We expect a model trained for natural language object retrieval to perform well when localizing a query relies on locating a specific object (e.g, a man in a brown shirt).  However, in general, the MCN model is able to retrieve correct moments more frequently than a model trained for natural language object retrieval.  DiDeMo is a difficult dataset and some queries, such as ``mother holds up green board for third time'' are not correctly localized by the MCN.}
\label{fig:nlor}
\end{figure*}

In the main paper, running a natural language object retrieval (NLOR) model on our data is a strong baseline. 
We expect this model to perform well on examples which require recognizing a specific object such as ``a man in a brown shirt runs by the camera'' (Figure\ref{fig:nlor} top row), but not as well for queries which require better understanding of action or camera movement such as ``man runs towards camera with baby'' (row 2 and 4 in Figure~\ref{fig:nlor}).
Though the Moment Context Network performs well on DiDeMo, there are a variety of difficult queries it fails to properly localize, such as ``Mother holds up the green board for the third time'' (Figure~\ref{fig:nlor} last row).

Please see {\tt\small \url{https://www.youtube.com/watch?v=MRO7_4ouNWU}} for examples of moments correctly retrieved by our model.

\section{Additional Baselines}

In the main paper we compare MCN to the natural language object retrieval model of \cite{hu2015natural}.
Since the publication of \cite{hu2015natural}, better natural language object retrieval models have been proposed (e.g., \cite{hu2016modeling}).
We evaluate \cite{hu2016modeling} on our data, in a similar way to how we evaluated \cite{hu2015natural} on our data in the main paper (Table 3 Row 5 in the main paper).
We extract frames at 10 fps on videos in our test set and use \cite{hu2016modeling} to score each bounding box in an image for our description.
The score for a frame is the max score of all bounding boxes in the frame, and the score for a moment is the average of all frames in the moment.
We expect this model to do well when the moment descriptions can be well localized by localizing specific objects.  
Surprisingly, even though CMN outperforms \cite{hu2015natural} for natural language object retrieval, it does worse than \cite{hu2015natural} on our data (Table~\ref{tab:results-supp} row 6).
One possible reason is that \cite{hu2016modeling} relies on parsing subject, relationship, and object triplets in sentences.
Sentences in DiDeMo may not fit this structure well, leading to a decrease in performance.
Additionally, \cite{hu2016modeling} is trained on MSCOCO~\cite{mscoco} and \cite{hu2015natural} is trained on ReferIt \cite{kazemzadeh2014referitgame}.
Though MSCOCO is larger than ReferIt, it is possible that the images in ReferIt are more similar to ours and thus \cite{hu2015natural} transfers better to our task.

Additionally, we train \cite{karpathy2014deep}, which is designed for natural language image retrieval, using our data.  
\cite{karpathy2014deep} relies on first running a dependency parser to extract sentence fragments linked in a dependency tree (e.g., ``black dog'', or ``run fast'').
It scores an image based on how well sentence fragments match a set of proposed bounding boxes.
To train this model for our task, we also extract sentence fragments, but then score temporal regions based on how well sentence fragments match a ground truth temporal region.  
We train on our data (using a late fusion approach to combine RGB and optical flow), and find that this baseline performs similarly to other baselines (Table~\ref{tab:results-supp} row 8).  In general, we believe our method works better than other baselines because it considers both positive and negative moments when learning to localize video moments and directly optimizes the R$@$1 metric.

\begin{table}[t]
\setlength{\tabcolsep}{5pt} %
\setlength\extrarowheight{1.8pt}
\centering
\small
\footnotesize
\rowcolors{3}{white}{blue!10}
\resizebox{\linewidth}{!}{%

  \begin{tabular}{llrrr}
    \toprule
     \multicolumn{5}{c}{\textbf{Baseline Comparison (Test Set)}}\\
     \midrule
     & Model & Rank$@$1 & Rank$@$5 & mIoU \\
    \midrule
    \midrule
     1 & Upper Bound & 74.75 & 100.00 & 96.05 \\    
     2 & Chance &  3.75 & 22.50 & 22.64 \\
     3 & Prior (tef) & 19.40 & 66.38 & 26.65 \\
     4 & CCA & 16.27 & 41.82 & 35.73 \\
     5 & Natural Lang. Obj. Retrieval (SCRC~\cite{hu2015natural}) & 16.20  & 43.94 & 27.18 \\
     6 & Natural Lang. Obj. Retrieval (CMN~\cite{hu2016modeling}) & 12.59  & 38.52 & 22.50 \\
     7 & Natural Lang. Obj. Retrieval (SCRC~\cite{hu2015natural} re-trained) & 15.57 & 48.32 & 30.55 \\
     8 & Image Retrieval (DeFrag~\cite{karpathy2014deep} re-trained) & 10.61 & 33.00 & 28.08 \\
    \midrule
    9  & \modelname{} (ours)  & \textbf{28.10} & \textbf{78.21} & \textbf{41.08} \\
    \bottomrule 
    \midrule
    \rowcolor{white}
    \multicolumn{5}{c}{\textbf{Ablations (Validation Set)}}\\
    10 & \modelname{}: Inter-Neg. Loss &  25.58 & 74.13 & 39.77 \\
    11 & \modelname{} Intra-Neg. Loss &  26.77 & 78.13 & 39.83\\
    12 & \modelname{} &  \textbf{27.57} & \textbf{79.69} & \textbf{41.70}\\
    \midrule
  
  \end{tabular}
  }
    \vspace{.3cm}
    \normalsize
  \caption{\modelname{} outperformes baselines (rows 1-8) on our test set.  We show ablation studies for our inter-intra negative loss in rows 10-12. 
}
  \label{tab:results-supp}
\end{table}

\section{Inter-Intra Negative Loss}

In Table~\ref{tab:results-supp} we compare results when training with only an inter-negative loss, only an intra-negative loss, and our proposed inter-intra negative loss.
Considering both types of negatives is important for best performance.

\section{Importance of Language Feature}
Because we ask annotators to mark any interesting moment and describe it, it is possible that annotators mark visually interesting moments which can be localized without text.  We thus train a model with our temporal context features but no text query and observe that this model outperforms chance and the moment frequency prior, but does not perform as well as our full model (25.04, 75.23, and 36.12 on R@1, R@5, and mIoU metrics).
This indicates that while understanding what constitutes a ``describable'' moment can be helpful for natural language moment retrieval, natural language is important to achieve best results on \datasetname{}.
Because the majority of videos include multiple distinct moments (86\%), we believe the gap between model trained with and without language will improve with better video-language modelling.

\section{Words Used to Construct Table 2}

To construct Table 2 in the main paper, we used the following words:

\begin{itemize}
\item Camera words: camera, cameras, zoom, zooms, pan, pans, focus, focuses, frame, cameraman
\item Temporal words: first, last, after, before, then, second, final, begin, again, return, third, ends
\item Spatial words:  left, right, top, bottom, background
\end{itemize}

Additionally, our vocab size is 7,785 words (which is large considering the total number of words in our dataset - 329,274).

\section{Video Retrieval Experiment}

\begin{figure*}[t]
\begin{center}
  \includegraphics[width=\linewidth]{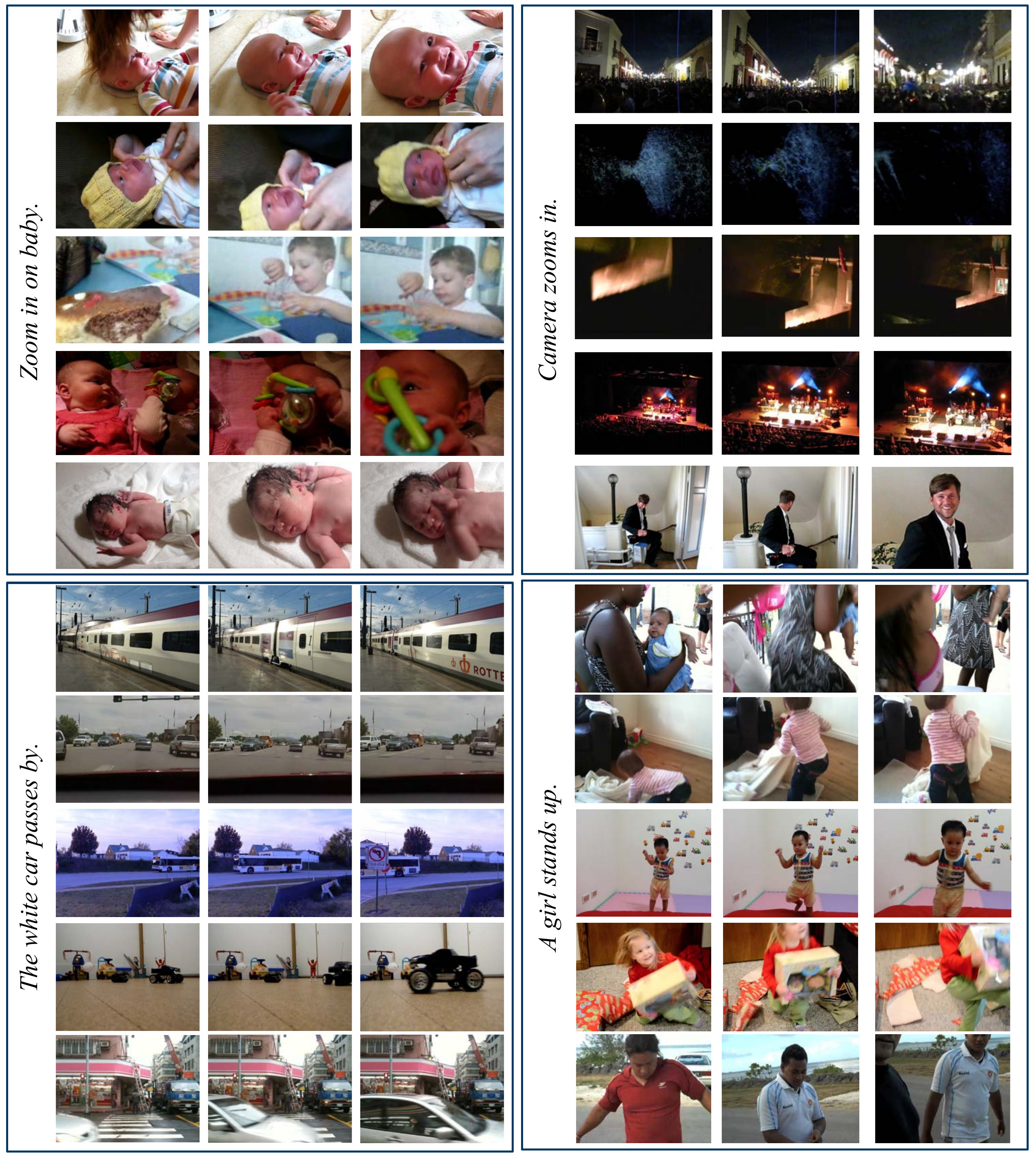}
 \end{center}
\caption{We use our model to retrieve the top moments which correspond to a specific query from the entire test set.  Though MCN was not trained to retrieve specific moments from a set of different videos, it is able to retrieve semantically meaningful results.  Above we show the top five moments retrieved for four separate text queries.  A video showing retrieved momenents can be found here: {\tt\small \url{https://www.youtube.com/watch?v=fuz-UBvgapk}}.}
\label{fig:moment_retrieval}
\end{figure*}

We used our model to retrieve five moments closest to a specific text query in our shared embedding space from all videos in our test set (Figure~\ref{fig:moment_retrieval}).
We find that retrieved moments are semantically similar to the provided text query.
For example, the query ``zoom in on baby'' returns moments in which the camera zooms in on babies or young children.
A similar query, ``camera zooms in'' returns example moments of the camera zooming, but the videos do not contain babies.
Though the query ``the white car passes by'' does not always return moments with cars, it returns moments which include semantically similar objects (trains, busses and cars).

Please see {\tt\small \url{https://www.youtube.com/watch?v=fuz-UBvgapk}} for an example of video retrieval results.

\section{Annotation Ambiguity}
\begin{figure*}[t]
\begin{center}
  \includegraphics[width=\linewidth]{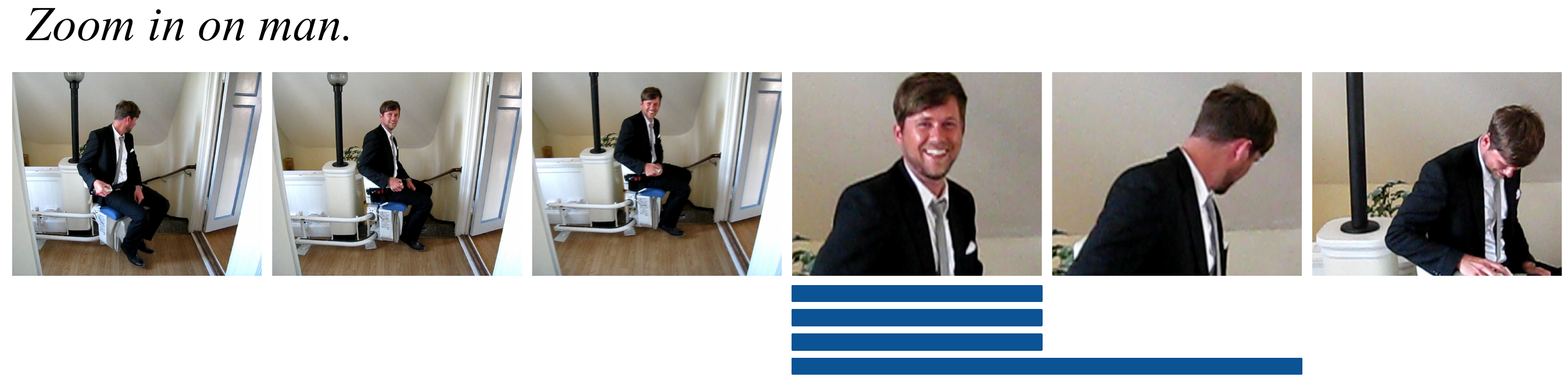}
 \end{center}
\caption{Humans do not always perfectly agree on start and end points for a moment.  In the above example we show annotations (denoted as blue lines) from four separate crowd-sourced annotators.  Though three annotators agree that the moment corresponds to the fourth segment, a fourth annotator believes the moment corresponds to both the fourth and fifth segment.  Our metrics reflect this ambiguity; a model which retrieves only the fourth segment will receive a high score.  A model which retrieves both the fourth and fifth segment will receive a lower score, but it will receive a higher score than a model which retrieves the third and fourth segments (which no annotators marked as the correct start and end point).}
\label{fig:metrics}
\end{figure*}

Figure~\ref{fig:metrics} shows an example in which the end point for specific moments are ambiguous.
For the query ``zoom in on man'', three annotators mark the fourth segment in which the camera actively zooms in on the man.
However, one annotator marks the segment in which the camera zooms in on the man and the following segment when the camera stays zoomed in on the man before zooming out.

This ambiguity informed how we chose our metrics.
Based on the annotations for the query ``zoom in on man'', it is clear that the moment retrieved by our model should include the fourth segment.
Though it is less clear if a moment retrieved by our model must include the fifth segment (which was only marked by one annotator to correspond to the phrase ``zoom in on man''), it is clear that a model which retrieves both the fourth and fifth segment is more correct than a model which retrieves the third and fourth segment.
When we compute a score for a specific example, we choose the maximum score when comparing the model's result to each four-choose-three combinations of human annotations.
This results in scores which reflect the intuition outlined above; a model which retrieves only the fourth segment (and therefore agrees with most annotators) will get a higher score than a model which retrieves the fourth and fifth segment (which only agrees with one annotator).
Additionally, a model which retrieves the fourth and fifth segment will receive a higher score than a model which retrieves the third and fourth segment.

Note that if two annotators had marked both the fourth and fifth segment, no retrieved moment would perfectly align with any four choose three combination of annotations.
Thus, for some examples, it is impossible for any model to achieve a perfect score.
In all our qualitative examples where we mark the ``ground truth'' moment in green, at least three annotators perfectly agree on the start and end point.

\section{Distribution of Annotated Moments}
\begin{figure*}[t]
\begin{center}
  \includegraphics[width=\linewidth]{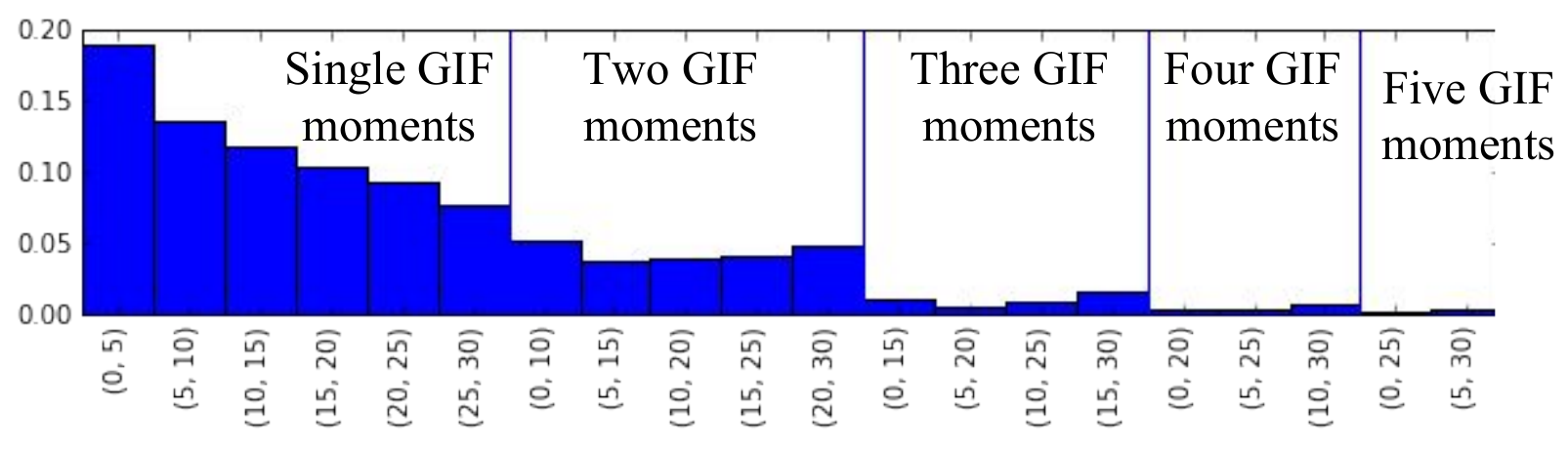}
 \end{center}
\caption{Distribution of segments marked in DiDeMo.  Moments tend to be short and occur towards the beginning of videos.}
\label{fig:gif_distribution}
\end{figure*}

Figure~\ref{fig:gif_distribution} shows the distribution of annotated start and end points in DiDeMo.
Moments marked by annotators tend to occur at the beginning of the videos and are short.
Though a ``prior baseline'' which retrieves moments which correspond to the most common start and end points in the dataset does much better than chance, our model significantly outperforms a ``prior baseline''.

\newpage

\newpage

\newpage

